\documentclass[10pt,twocolumn,letterpaper]{article}

 \usepackage{cvpr}              
\usepackage{array}
\usepackage{graphicx}

\usepackage[dvipsnames]{xcolor}

\definecolor{cvprblue}{rgb}{0.21,0.49,0.74}
\usepackage[pagebackref,breaklinks,colorlinks,citecolor=cvprblue]{hyperref}

\def\paperID{406} 
\def\confName{3DV\xspace}
\def\confYear{2025\xspace}

\title{DC3DO: Diffusion Classifier for 3D Objects}

\author{
\makebox[\textwidth][c]{
\begin{tabular}{c}
\textbf{Nursena Koprucu}\textsuperscript{1,*}, \textbf{Meher Shashwat Nigam}\textsuperscript{2,*}, \textbf{Shicheng (Luke) Xu}\textsuperscript{5,*}, \textbf{Biruk Abere,*}\textsuperscript{3}, \\
\textbf{Gabriele Dominici,*}\textsuperscript{4},
\textbf{Andrew Rodriguez}\textsuperscript{2}, 
\textbf{Sharvaree Vadgama}\textsuperscript{6}, 
\textbf{Berfin Inal}\textsuperscript{6}, 
\textbf{Alberto Tono}\textsuperscript{7,8}\\
\textsuperscript{1}Max Planck Institute for Intelligent Systems, Germany, \textsuperscript{2}Georgia Tech, USA,
\textsuperscript{3}University of Gondar, Ethiopia\\
\textsuperscript{4}Università della Svizzera Italiana, Switzerland,
\textsuperscript{5}Google LLC, USA,
\textsuperscript{6}University of Amsterdam, Netherlands\\
\textsuperscript{7}Stanford University, USA,
\textsuperscript{8}Computational Design Institute, USA\\
\end{tabular}
}
}

\begin{document}
\maketitle


\begin{abstract}

Inspired by Geoffrey Hinton’s emphasis on generative modeling (“To recognize shapes, first learn to generate them"), we explore the use of 3D diffusion models for object classification. Leveraging the density estimates from these models, our approach, “Diffusion Classifier for 3D Objects”, dubbed DC3DO, enables zero-shot classification of 3D shapes without additional training. Our method achieves on average 12.5\% improvement compared with its multi-view counterparts, demonstrating superior multimodal reasoning compared to discriminative approaches. DC3DO uses a class-conditional diffusion model trained on ShapeNet. We run inferences on chairs and cars pointclouds. This work underscores the potential of generative models in 3D object classification. Code available \href{https://github.com/SGI-2023/3D-Building-Classification}{https://github.com/SGI-2023/3D-Building-Classification}.

\end{abstract}

\section{Introduction}

Recent advancements in deep generative models have yielded \textit{state-of-the-art} (SOTA) performance in both classification and out-of-distribution (OOD) classification for images \cite{diffusion_classifier}. Deep generative models are increasingly being utilized for discriminative tasks, demonstrating superior effectiveness across various domains, including images \cite{huang2024activegenerationimageclassification}, text \cite{card}, and tabular data \cite{star_diffusion_models, huang2023dreamcontrol}. This progression builds upon the foundational work of Hinton \cite{recognizegeneratehinton07}, inspired by Oliver Selfridge’s “Pandemonium” model \cite{pandemoniumselfridge58}. While these early researchers focused on generation within the image domain, one could argue that creation should originate in 3D space before extending to the image or text domains. By first generating 3D objects, we can enhance downstream tasks, not only in object classification but also in image and text classification. 

\begin{figure}[ht]
    \centering
    \includegraphics[width=\columnwidth]{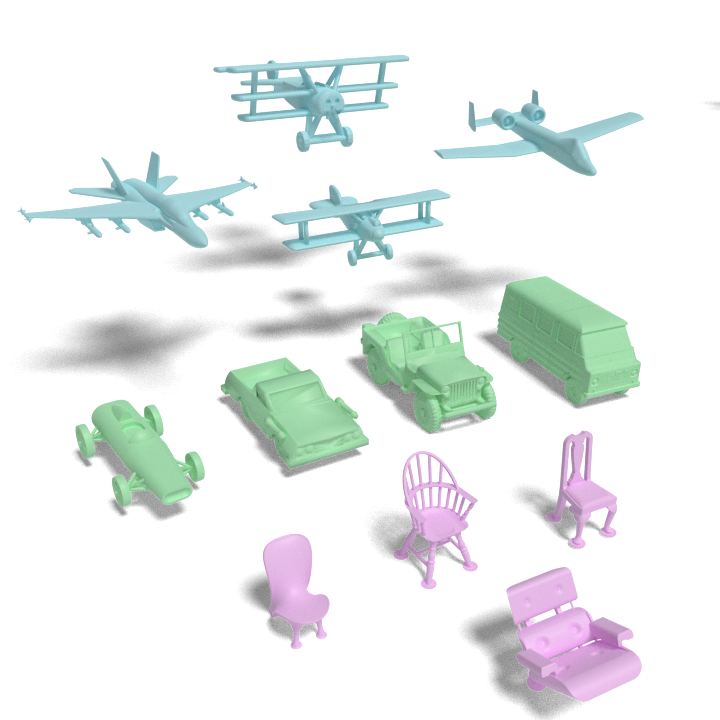} %
    \caption{ \textbf{Dataset classes for classification}. We performed 3D Classifications tests on cars, chairs, and airplanes. We used multi-view and point cloud representations only for chairs and cars.}
    \label{fig:hero}
\end{figure}

The classification of 3D shapes is increasingly important in fields such as computer vision, robotics, and virtual reality, hence this research. Since traditional methods often struggle to handle the complexity and variability inherent in 3D data, we adopted a diffusion approach \cite{ho2020denoisingdiffusionprobabilisticmodels}. Diffusion models \cite{SohlDicksteinW15diffusionunsupervised}, a recent class of likelihood-based generative models, have shown significant promise in various tasks~\cite{ramesh2022hierarchicaltextconditionalimagegeneration, ho2022imagenvideohighdefinition, poole2022dreamfusiontextto3dusing2d} by transforming random noise into coherent data samples through an iterative noising and denoising process.

Furthermore, today's work in diffusion models \cite{diffbetterthanGAN, diffae, DiffCLIP_Shen_2024_WACV} showed unmatched results not only on generative tasks \cite{pointclouddiffusion} but also in classification tasks \cite{meng2023concretescorematchinggeneralized, lou2024discretediffusionmodelingestimating}. Diffusion models belong to a class of generative models that model the data distribution of the dataset, similar to VAEs \cite{vae}, GANs \cite{3dgan, pigan}, EBMs \cite{xie2021generativepointnetdeepenergybased}, Score-based models \cite{yang2019pointflow3dpointcloud}. 

Therefore, a question arises, \textit{Can we use diffusion models for 3D classification tasks?} More critically, given their remarkable ability to generate \textit{original} objects  beyond the initial dataset distribution \cite{lion, nam20223dldm, TONOVitruvio22, pointclouddiffusion, zheng2023lasdiffusion}, how do these models perform for out-of-distribution (OOD) data. While these models have excelled on standard benchmarks, they often struggle with novel OOD data, a limitation attributed to biased training datasets that fail to encompass the full spectrum of real-world possibilities. This has been attributed to the biased training data that does not represent all real-world possibilities \cite{jahanian2020steerabilitygenerativeadversarialnetworks}. These deep generative models can synthesize strikingly realistic and diverse images, objects, and text and they have shown better performances in zero-shot \cite{jain2021zeroshot, dreamfields_zeroshot, sanghi2023sketchashapezeroshotsketchto3dshape}, few-shot classification tasks \cite{DiffCLIP_Shen_2024_WACV}. 

\label{sec:intro}


In this research, we explore the application of Denoising Diffusion Probabilistic Models (DDPMs)~\cite{diffusion_classifier} for classifying 3D shapes. Traditional classification methods often fall short with 3D data, requiring novel approaches. Furthermore, 3D data are represented as point clouds \cite{pointbert, pointclouddiffusion, pointvoxeldiffusion}, voxels \cite{3dr2n2, marrnet}, signed distance functions \cite{TONOVitruvio22, nam20223dldm, zeng2022lion} and multi-view formats \cite{su15mvcnn}. Inspired by LION \cite{lion}, this research adopts point cloud and voxel \cite{pointvoxeldiffusion, lion} combined with latent representations \cite{nam20223dldm} and diffusion models: DC3DO. DC3DO focuses on leveraging the generative capabilities of diffusion models for zero-shot classification~\cite{li2023diffusionmodelsecretlyzeroshot}. We compared it against a direct extenstion of the 2D conterpart performed on images \cite{diffusion_classifier}. In a dynamic data landscape, the ability to classify data into previously unseen categories, such as architectural structures~\cite{Encodedmemory, componet2021, federova}, is of paramount importance. Diffusion models, with their inherent generative strengths, are particularly well-suited for this challenge. Furthermore, by advancing beyond traditional 2D prior models~\cite{liu2023one2345singleimage3d, liu2023one2345++, liu2023zero1to3} and incorporating the LION model~\cite{lion}, renowned for generating high-fidelity 3D shapes, we enhance the effectiveness of the \textit{Diffusion Classifier} in performing discriminative tasks, particularly in 3D object classification. Therefore, our contributions to this field are three-fold:

\begin{itemize}
    \item \textbf{Novel Method for 3D shape classification:} We introduce DC3DO to classify 3D shapes with a diffusion model.

    \item \textbf{Comparative analysis:} We compare our method against multiview 3D representations using a 2D diffusion classifier \cite{diffusion_classifier}. We adapted MVCNN's~\cite{su15mvcnn} with its view pooling method to a more U-Net and diffusion classifier-friendly method for a fair comparison.

\end{itemize}

In these unsupervised settings, diffusion models \cite{ho2020denoisingdiffusionprobabilisticmodels} are trained using the objectives of Variational Inference, specifically focusing on maximizing the evidence lower bound (ELBO)~\cite{sutter2021generalizedmultimodalelbo} of the log-likelihood, as described in \cite{diffusion_classifier}. This involves adding noise $\epsilon$ to a sample, using a neural network to predict the noise, and adopting Mean Squared Error (MSE) or L2 loss to compare this predicted noise against a white gaussian noise \cite{Bluegaussiannoise24}. 

Given an input $\mathbf{x}$ and a finite set of classes $\mathbf{c}$ that we want to choose from, we can use the model to compute class-conditional likelihoods $p_\theta(\mathbf{x} \mid \mathbf{c_i})$, where $i$ indicated the class number. Then, by selecting an appropriate prior distribution $p(\mathbf{c_i})$ and applying Bayes' theorem, we can obtain predicted class probabilities $p(\mathbf{c_i} \mid \mathbf{x})$. For conditional diffusion models that use an auxiliary input, like a class index for class-conditioned models or a prompt for text-to-image models, we can achieve this by leveraging the ELBO as an approximate class-conditional log-likelihood $\log p(\mathbf{x} \mid \mathbf{c_i})$. We repeatedly add noise and compute an estimate of the expected noise reconstruction losses (also called $\epsilon$-prediction loss) for every class, as described in \cite{diffusion_classifier}, please refer to Figure \ref{fig:LIONdiffusion}.

\section{Related Work}

Multimodal large language models (LLMs) strengths are leveraged in many current works \cite{qi2024shapellmuniversal3dobject, ji2024jm3djm3dllmelevating, xu2023pointllmempoweringlargelanguage, guo2023pointbindpointllmaligning}. LLMs can handle diverse tasks through conversational interaction, specifically in the context of 3D objects. Typically, this is achieved by training a 3D shape encoder and aligning it with other modalities (e.g., text, images, and audio). The entire pipeline is then fine-tuned during an instruction-tuning phase, resulting in a model that is better aligned with user requests for specific 3D tasks. This fine-tuning stage is conducted using synthetic datasets or captioning datasets. 
These approaches highlight the vast potential of integrating 3D shapes into foundation models, although they still necessitate the fine-tuning of large models. Other methods, such as 3DAxiesPrompts \cite{liu20233daxiespromptsunleashing3dspatial}, enhance images and prompts with additional artifacts to be able to exploit the 2D vision abilities of LLM for 3D objects.

PEVA-Net \cite{lin2024pevanetpromptenhancedviewaggregation} employs a pre-trained CLIP model in a multiview pipeline to classify 3D objects in zero-shot or few-shot environments. It leverages CLIP’s zero-shot classification abilities for each view of the 3D object, subsequently aggregating these results to make the final prediction. Although this approach effectively exploits the zero-shot capabilities of vision-language models (VLMs), transforming 3D shapes into multiview images is an oversimplification that can lead to suboptimal results.

TAMM \cite{zhang2024tammtriadaptermultimodallearning} demonstrates that when aligning 3D object representations with other modalities, the image modality contributes less significantly than the text modality. To address this, their method learns to separate visual features from semantic features within the 3D object representation, enabling a more effective alignment with the other modalities and enhancing performance in downstream tasks. These findings suggest that the alignment between modalities for integrating 3D representations into existing methods can sometimes be inadequate \cite{Ulip_2_Xue2024}. Regarding 3D representation learning, Zhang et al. \cite{learning3drepresentations2d} takes a different approach and incorporates 2D guidance. Their work, dubbed I2P-MAE \cite{learning3drepresentations2d}, learns advanced 3D representations, achieving state-of-the-art performance on 3D tasks and significantly lowering the need for large-scale 3D datasets. On the contrary concurrent work, DiffCLIP \cite{DiffCLIP_Shen_2024_WACV} demonstrates that the integration of CLIP and diffusion models for 3D classification facilitates zero-shot classification, achieving state-of-the-art results. This methodology utilizes a pre-training pipeline that incorporates a Point Transformer for few-shot 3D point cloud classification, wherein the CLIP model extracts style-based features of the class, synergistically combined with image features. While DiffCLIP \cite{DiffCLIP_Shen_2024_WACV} used Point Transformer we used LION, a latent point-voxel \cite{liu2019pvcnn, pointvoxeldiffusion, lion} representations that leverages a hierarchical two stages diffusion process with state of the art generative performances. Following the line of latent and implicit representations, Xin et al. \cite{text23dclassifier23} used a Classifier Score Distillation (CSD) method, which utilizes an implicit classification model for generation.

\section{Methodology}

In this section, we present and compare two distinct approaches for 3D object classification: Multi-View Diffusion Classifier (Section \ref{subsec:multi_view_diffusion}) and DC3DO (Section \ref{subsec:lion_diffusion}). The first approach, the Multi-View Diffusion Classifier, is designed to harness the power of diffusion models while maintaining the architecture of \textit{Diffusion Classifier} \cite{diffusion_classifier}. The second approach, DC3DO, integrates the advanced generative capabilities of LION \cite{lion} with diffusion-based classification, targeting zero-shot classification of complex 3D shapes like cars and chairs.

Our goal is to thoroughly assess the performance of these methods in comparison to traditional and state-of-the-art techniques. The Multi-View Diffusion Classifier (MVDC) offers an alternative to the widely-used MVCNN by employing a majority vote mechanism across multiple 3D views. On the other hand, DC3DO leverages LION’s robust generative framework (Section \ref{subsec:integration_lion_diffusion}).

\subsection{Multi-View Diffusion Classifier (MVDC)}
\label{subsec:multi_view_diffusion}

3D objects can be effectively represented as a series of images, providing a straightforward baseline for extending previous work \cite{diffusion_classifier} to the 3D domain. By simply aggregating multiple views of the same object, we can adapt existing diffusion-based classification techniques for 3D shapes. For our experiments, we utilized the ShapeNet dataset \cite{chang2015shapenet}, focusing on a specific subset of $200$ models per class. This selection was made due to the computational intensity of performing $1000$ diffusion steps ($t$) per image ($X_i$), which is particularly challenging in environments with limited GPU resources (\textit{poor-gpus settings}). Especially if each object is represented by $36$ views taken from cameras view 10-degree intervals around a circumference encircling the 3D object, adding to the computational complexity. 

To classify a single object, our method proposes a majority vote scheme. Let $X=\{X_1, X_2, \ldots, X_n\}$ represent the set of $n$ views of a 3D shape. Each view $X_i$ is processed individually by the diffusion model to produce a corresponding prediction $y_i=f\left(X_i\right)$, where $f(\cdot)$ denotes the classification function of the diffusion model. Unlike MVCNN \cite{su15mvcnn}, which aggregates these views into a global representation through view pooling, our approach maintains the predictions $\left\{y_1, y_2, \ldots, y_n\right\}$ independently.

The final classification decision $y^*$ is made by selecting the most common prediction among the individual predictions, formulated as:
$$
y^*=\operatorname{mode}\left(y_1, y_2, \ldots, y_n\right)
$$

This majority vote approach retains the architectural integrity of diffusion models while emphasizing simplicity and interpretability. While MVCNN's view pooling may enhance performance by combining features, our goal is to assess the classification power of diffusion models in their unaltered form. Here the 2D Images are processed and encoded into feature maps $\mathbf{i} \in \mathbb{R}^{H \times W \times C}$, where $H$ is the height, $W$ is the width, and $C$ is the number of channels in the image ($512 \times 512 \times 3$ as the highest resolution in these experiments).

\subsection{Diffusion Classifier for 3D Objects (DC3DO) }
\label{subsec:lion_diffusion}

DC3DO consists on the main contribution of our research. Our model combines LION \cite{lion} with diffusion classifier \cite{diffusion_classifier} for zero-shot classification. By utilizing LION's ability to generate diverse 3D shapes and feeding them into the diffusion classifier, we achieve precise categorization of 3D cars and chairs. This section details DC3DO, emphasizing the advantages of diffusion models, the processing of 3D shapes, and the integration of the LION model to enhance classification accuracy.

\subsection{Integrating LION with Diffusion Classifiers}
\label{subsec:integration_lion_diffusion}
\begin{figure*}[h!]
    \centering
    \includegraphics[width=\textwidth]{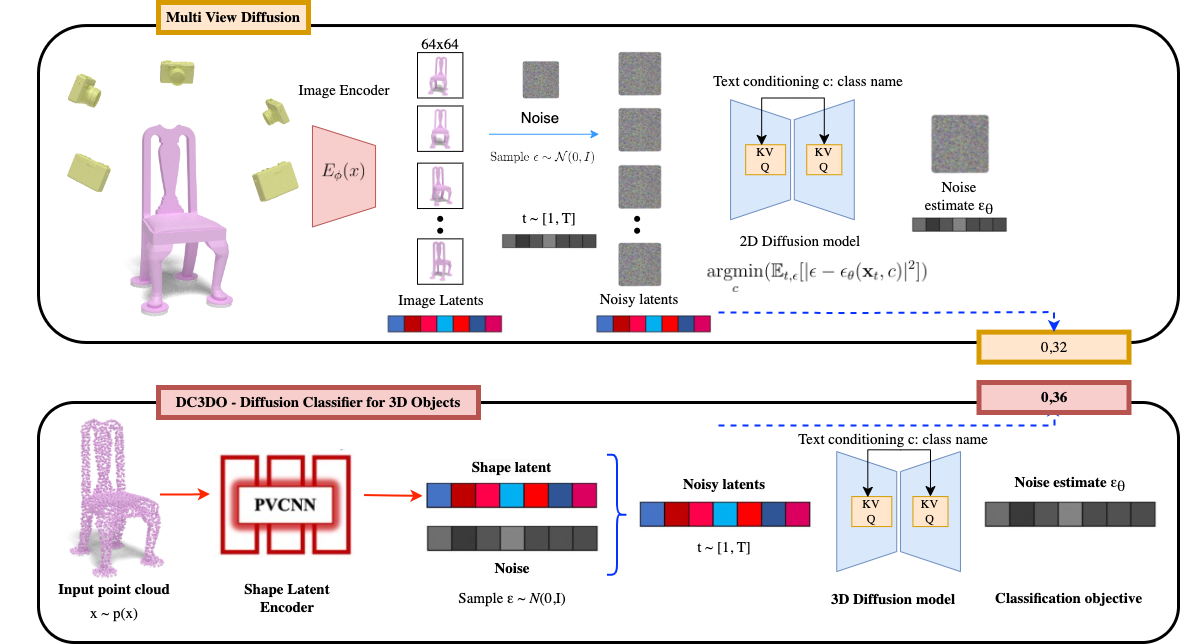} 
    \caption{ \textbf{Methods comparison}. We extended the \textit{Diffusion Classifier} \cite{diffusion_classifier} paper to a multi-view \cite{su15mvcnn} settings and we compare with our DC3DO model, based on \cite{lion}}
    \label{fig:LIONdiffusion}
\end{figure*}

LION leverages a hierarchical latent space to effectively capture both global and local features of 3D structures, see Figure \ref{fig:LIONdiffusion}. This hierarchical approach ensures comprehensive encoding of both macro and micro features of 3D object structures. 

\subsubsection{Integration with Hierarchical Latent Space}
LION's hierarchical latent space encodes 3D point clouds $\mathbf{x} \in \mathbb{R}^{3 \times N}$, where $\mathbf{x}$ consists of $N$ points (2048) with $xyz$-coordinates in $\mathbb{R}^3$, into a dual-layered latent representation. This representation includes:

\begin{itemize}
    \item \textbf{Global Latent Space}: This vector-valued latent space, denoted as $\mathbf{z}_0 \in \mathbb{R}^{D_{\mathbf{z}}}$, captures the overall structure and large-scale features of the building. It captures the overall spatial structure of the 3D shape, including its large-scale features.
    \item \textbf{Local Point-Structured Latent Space}: This latent space, denoted as $\mathbf{h}_0 \in \mathbb{R}^{(3 + D_{\mathbf{h}}) \times N}$, represents a point cloud-structured latent consisting of $N$ points with $xyz$-coordinates in $\mathbb{R}^3$ and additional $D_{\mathbf{h}}$ latent features per point. This layer captures detailed and fine-grained features. 
\end{itemize}

\subsubsection{Integration with Diffusion Models}

The integration process involves several key steps:

\paragraph{Encoding}
The 3D point cloud data $\mathbf{x}$ is encoded into the global latent space using LION's PVCNN encoder. We found that the global latents contained enough information about the shape and high level features of the object, for the purpose of classification. Also, it is a much smaller latent space compared to the the local point structured latent space, making it easier to work with - considering the diffusion process requires multiple inference steps per sample. 

\paragraph{Diffusion Process}
After encoding, the data undergoes a first diffusion process (global latent). This involves a fixed forward procedure where Gaussian noise is iteratively added (1000 steps) to the latent representations $\mathbf{z}_0$ and $\mathbf{h}_0$, resulting in the diffused latents $\mathbf{z}_t$ and $\mathbf{h}_t$. The forward diffusion process is defined as:
\begin{align}
    \mathbf{z}_t = \alpha_t \mathbf{z}_0 + \sigma_t \mathbf{\epsilon}, \quad \mathbf{\epsilon} \sim \mathcal{N}(\mathbf{0}, \mathbf{I})
\end{align}

\begin{align}
    \mathbf{h}_t = \alpha_t \mathbf{h}_0 + \sigma_t \mathbf{\epsilon}, \quad \mathbf{\epsilon} \sim \mathcal{N}(\mathbf{0}, \mathbf{I})
\end{align}

where:
\begin{itemize}
    \item $\mathbf{z}_0$ and $\mathbf{h}_0$ are the initial latent representations capturing the global and local features of the 3D shape, respectively.
    \item $\mathbf{z}_t$ and $\mathbf{h}_t$ are the diffused latent representations at timestep $t$.
    \item $\alpha_t$ and $\sigma_t$ are coefficients that control the amount of the original signal and the noise added at each timestep, respectively.
    \item $\mathbf{\epsilon}$ is the Gaussian noise sampled from a standard normal distribution $\mathcal{N}(\mathbf{0}, \mathbf{I})$, which introduces randomness to the latent representations.
\end{itemize}

\paragraph{Denoising and Classification}
In the pipeline, a deep neural network, conditioned on class labels $\mathbf{c}$, performs the denoising of the perturbed data. The denoising process involves reversing the forward diffusion to retrieve the latent representations $\hat{\mathbf{z}}_0$ and $\hat{\mathbf{h}}_0$ that best match the original data distribution $\mathbf{x}$. Specifically, the network learns to approximate the posterior distributions $q_\phi(\mathbf{z}_0|\mathbf{x}, \mathbf{c})$ and $q_\phi(\mathbf{h}_0|\mathbf{x}, \mathbf{z}_0, \mathbf{c})$ by minimizing the reconstruction error.

The classification is then performed by evaluating the likelihood $p_\theta(\mathbf{x}_0 \mid \mathbf{c})$ of the denoised data $\hat{\mathbf{z}}_0$ and $\hat{\mathbf{h}}_0$ belonging to specific classes. This likelihood is computed as follows:
\begin{align}
    p_\theta(\mathbf{x}_0 \mid \mathbf{c}) = \int_{\mathbf{x}_{1:T}} p(\mathbf{x}_T) \prod_{t=1}^T p_\theta(\mathbf{x}_{t-1} \mid \mathbf{x}_t, \mathbf{c})\ \mathrm{d}\mathbf{x}_{1:T}
\end{align}

where
\begin{itemize}
    \item $p_\theta(\mathbf{x}_0 \mid \mathbf{c})$ is the class-conditional likelihood of the original data $\mathbf{x}_0$ given the class label $\mathbf{c}$.
    \item $\mathbf{x}_T$ represents the final diffused state, typically modeled as a standard Gaussian distribution.
    \item $p_\theta(\mathbf{x}_{t-1} \mid \mathbf{x}_t, \mathbf{c})$ denotes the learned reverse process that denoises the data at each timestep $t$, conditioned on the class label $\mathbf{c}$.
\end{itemize}

The model assesses the denoised data to determine the most likely building category by evaluating which class-specific denoising process best corresponds to the introduced noise, ultimately assigning the data point to the class with the highest likelihood.

\subsubsection{Text-Conditioned Diffusion}

Our model employs multi-modal \cite{xu2023multitask3dbuildingunderstanding} approach to integrate diverse data modalities, providing a comprehensive approach to 3D building classification. The diffusion process is condition on a text prompt \textit{"[C]"} ("car", "chair", and  "airplane" ). We added an additional text prompt to provide a diffusion process to classify the model as "non-car", "non-chair", and "non-airplane". 

The integrated modalities include:

\begin{itemize}
    \item \textbf{3D Point Cloud Data}: The primary representation of 3D shapes, capturing spatial distribution and structural details.
    \item \textbf{Textual Descriptions}: Supplementary information describing the architectural features and styles of buildings. These are encoded into vector representations $\mathbf{t} \in \mathbb{R}^d$.
\end{itemize}

By integrating these modalities, our model achieves a more informed representation of the data, which improves classification accuracy and robustness. The integration of LION with diffusion models utilizes the combined strengths of both techniques, allowing for precise and reliable classification of 3D building structures.

\section{Experimental Results}

\subsection{MVDC - 2D Results}

In our baseline evaluation, we utilized a multi-view diffusion classifier on the ShapeNet dataset, focusing on three classes: cars, chairs, and airplanes. This approach uses multiple views of 3D shapes to enhance classification accuracy by taking advantage of the rich spatial information in the dataset. The process involved encoding 3D shapes into latent representations using a pretrained VAE, adding Gaussian noise, and employing a UNet model for denoising and classification. This way, we captured the intricate details of 3D shapes and effectively categorized them by adaptively selecting the most promising samples based on predicted errors, optimizing overall classification performance.


\begin{table}[h]
    \centering
    \caption{Zero-shot classification accuracy (\%) of DC3DO on ShapeNet for cars, airplanes, and chairs. We performed the comparison only on the first $200$ subsamples models, each with 6 views and 200 sampling steps. For the Multi View Diffusion we used only the six frontal views, and image resolutions of $64 \times 64$.}
    \begin{tabular}{l|*{3}{c}}
        \toprule
        \textbf{Method} & \multicolumn{2}{c}{\textbf{Accuracy}} \\
        \cmidrule(lr){2-3}
        & \textbf{Car} & \textbf{Chair}  \\
        \midrule
        MVDC (100 models) & 65.7\% & 32.3\% \\
        MVDC (200 models) & 64.8\% & 31.5\%  \\
        \midrule
        \textbf{DC3DO-100m (ours)} & \textbf{100\%} & \textbf{36\%}  \\
        \textbf{DC3DO-200m (ours)} & \textbf{100\%} & \textbf{49\%}  \\
        \bottomrule
    \end{tabular}
\end{table}

MVCNN \cite{su15mvcnn} utilized 36 fixed cameras, with objects placed in a canonical pose. The cameras were positioned at uniform intervals, with a 10-degree rotation between each, defined by their position parameters $(X, Y, Z)$. To manage computational constraints, we downscaled the images to $64 \times 64$ and reduced the number of views per object from 36 to 6. The selected views were the frontal ones, corresponding to camera angles of $10^{\circ}, 20^{\circ}, 30^{\circ}, 340^{\circ}, 350^{\circ}$, and $360^{\circ}$, in line with the ShapeNet view settings.

As explained in \cite{DiffCLIP_Shen_2024_WACV}, frontal camera positions generally yield higher accuracy. Therefore, we focused on these 6 specific camera positions for our experiments.

To compute the \textbf{accuracy} of multi-view classification, we employed a majority vote mechanism across the $n$ views of each 3D mesh, where $n=6$. Let $y_i \in\{0,1\}$ represent the binary prediction for each view $X_i$, where 1 corresponds to the prediction ``car'' and 0 corresponds to ``not car''. The final classification $y^*$ for each object is determined as:
$$
y^*= \begin{cases}1 & \text { if } \sum_{i=1}^n y_i \geq \frac{n}{2} \\ 0 & \text { otherwise. }\end{cases}
$$

In this setup, if the number of votes for "car" (i.e., $\sum_{i=1}^n y_i$ ) is greater than or equal to 3, the object is classified as "car." For each class $c$, we performed a binary classification, distinguishing between class $c$ and all other classes. The accuracy $A_c$ for each class $c$ is calculated as:
$$
A_c=\frac{\text { Number of correctly classified objects in class } c}{\text { Total number of objects in class } c} .
$$

Finally, the mean per-class accuracy $\bar{A}$ is computed by averaging the binary classification accuracies across all classes:
$$
\bar{A}=\frac{1}{C} \sum_{c=1}^C A_c,
$$
where $C$ is the total number of classes.

\subsection{DC3DO Inference}

Since DC3DO is based on LION, we took its model weights publicly available. LION has been trained on specific classes from the ShapeNet dataset, with the model weights for the “chairs” and “cars” categories publicly available to the research community. Consequently, we utilized these pretrained models in our experiments. Due to computational constraints, we set the number of diffusion steps for both the Multi-View Diffusion Classifier and LION to 200 steps.

For our experiments, we employed a batch size of $1$ for the “cars”  and “chairs” categories. Currently, it takes approximately 20 seconds to classify each object. 











\begin{table}[ht]
\centering
\begin{tabular}{c|c|c}
\textbf{Model}     & \textbf{Render} & \textbf{Accuracy}        \\ \hline
\textit{Chair 142} &   {\includegraphics[width=2.5cm]{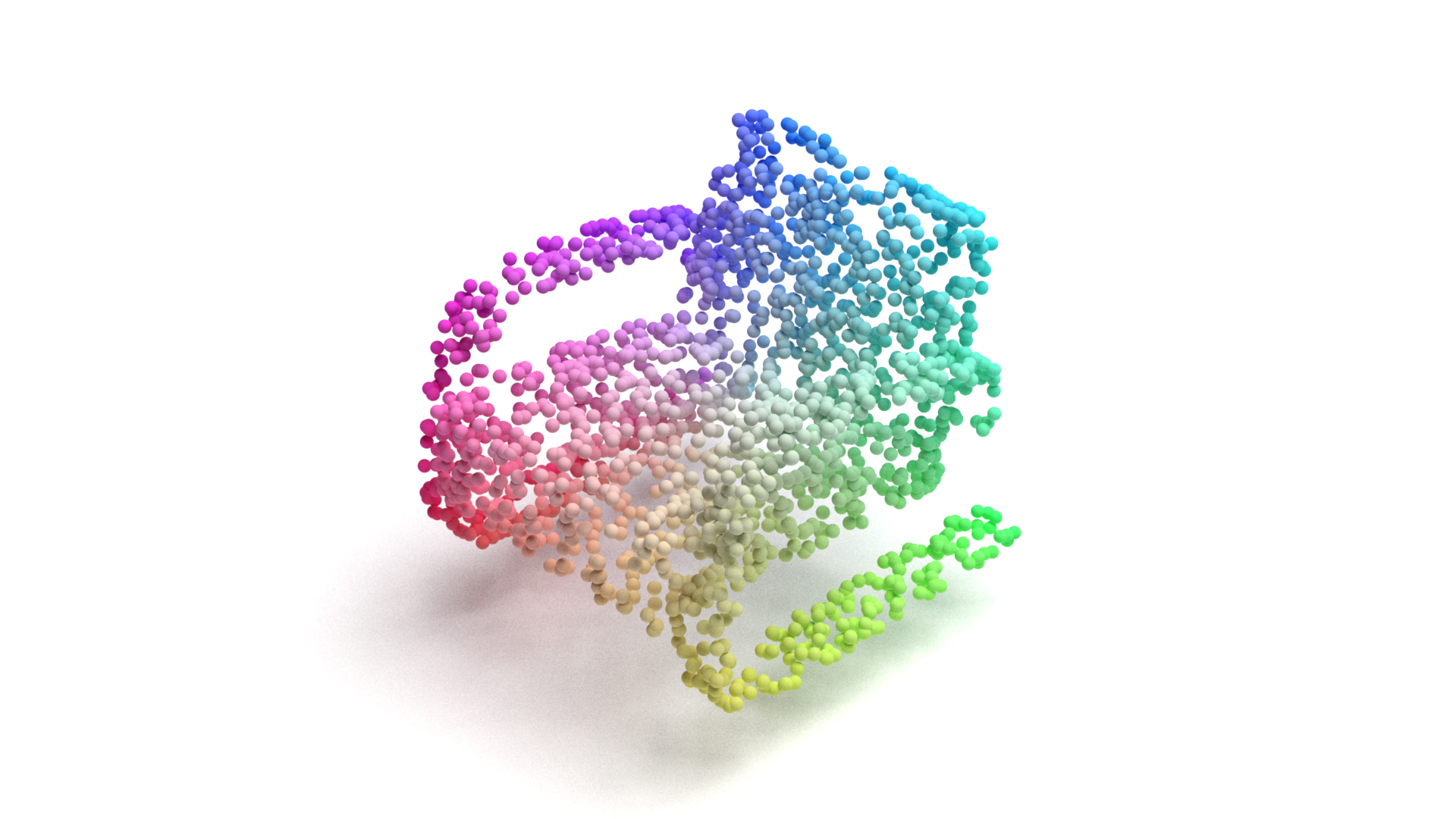}}           & 35\%    \\
\textit{Chair 143} & \includegraphics[width=2.5cm]{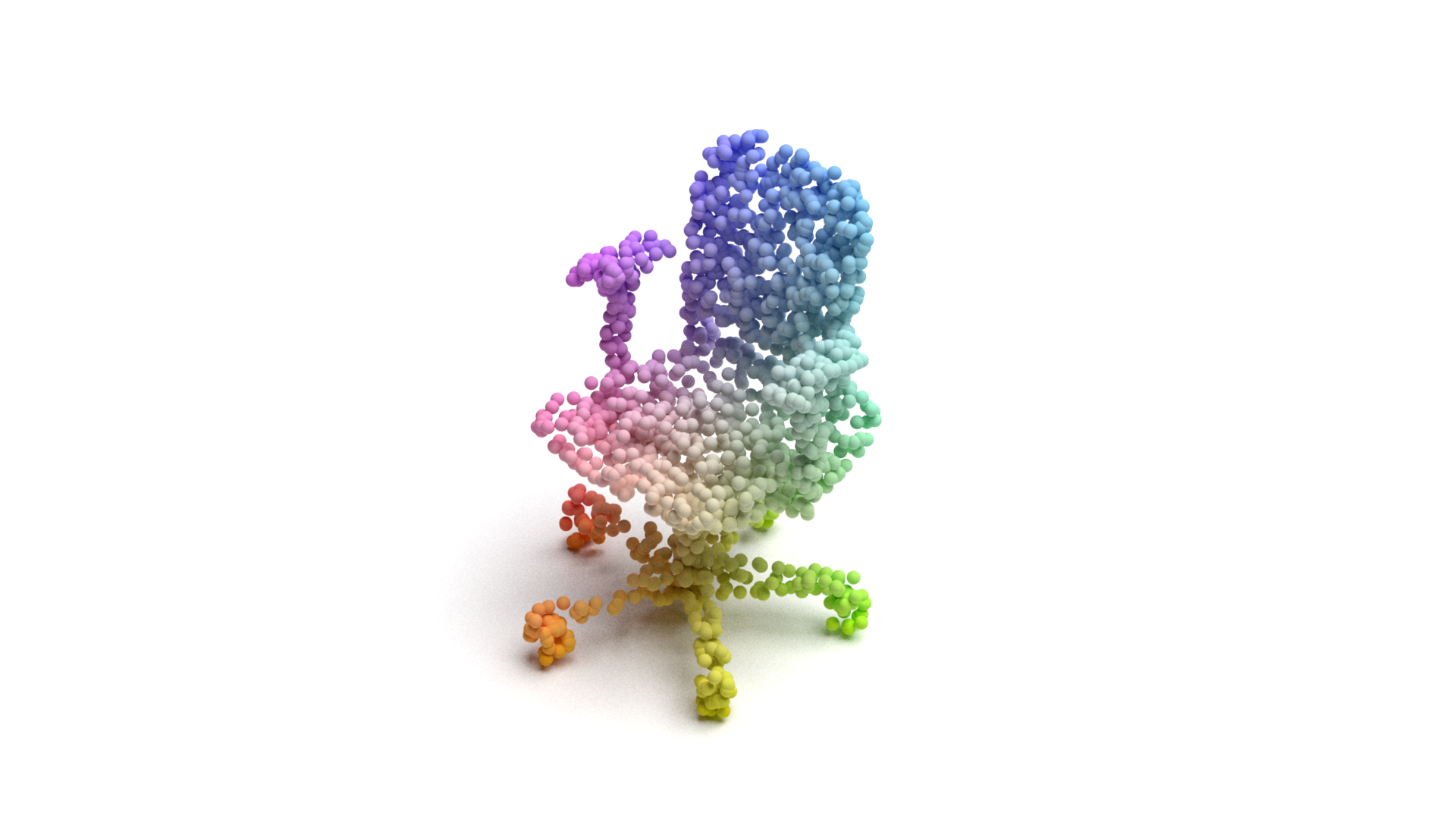}                &    34\%                       \\
\textit{Chair 188} & \includegraphics[width=2.5cm]{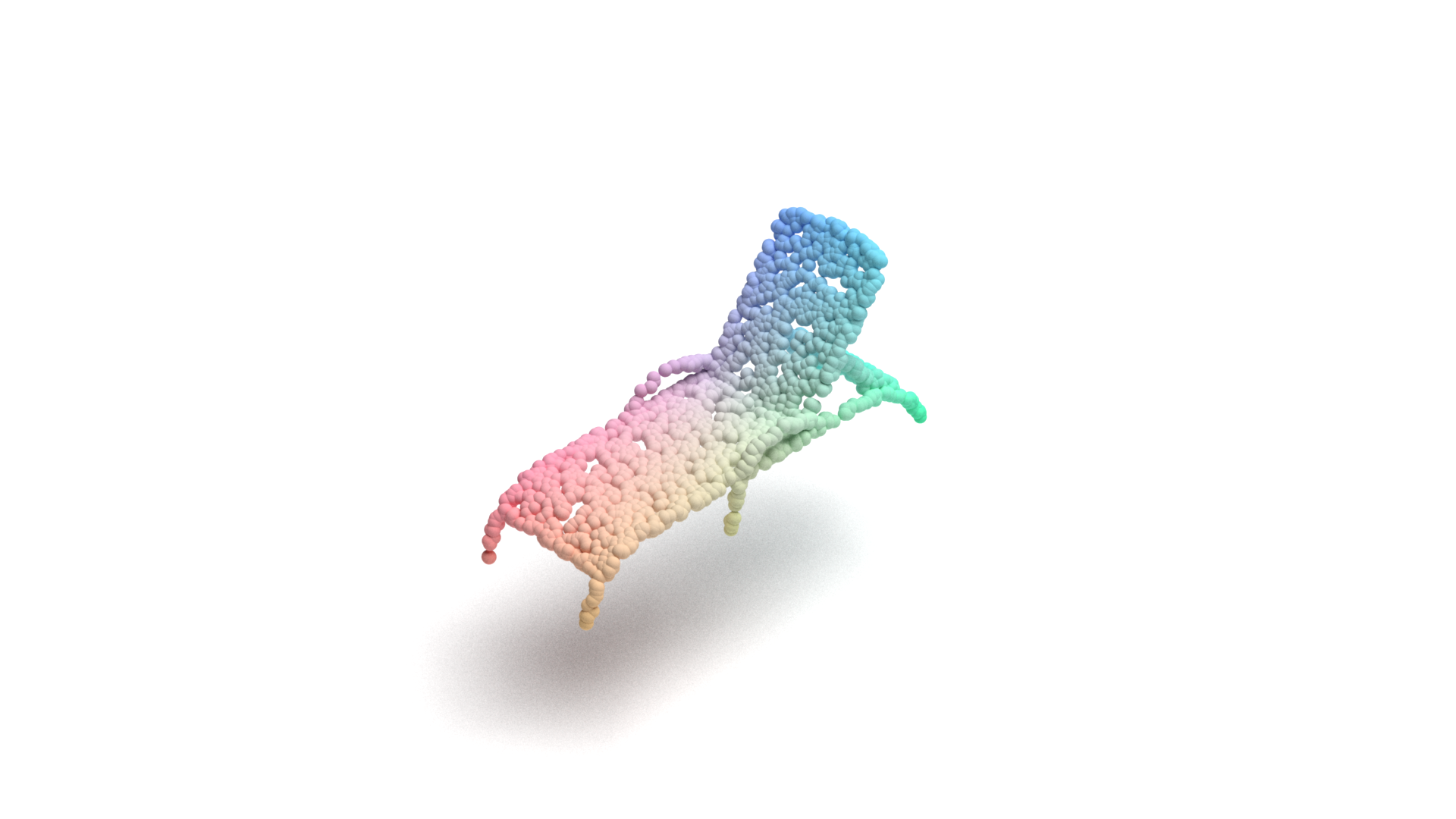}                &     33\%                     
\end{tabular}
\caption{Best 3 chairs per accuracy}
\end{table}

\begin{table}[ht]
\centering
\begin{tabular}{c|c|c}
\textbf{Model}     & \textbf{Render} & \textbf{Accuracy}        \\ \hline
\textit{Chair 116} &   {\includegraphics[width=2.5cm]{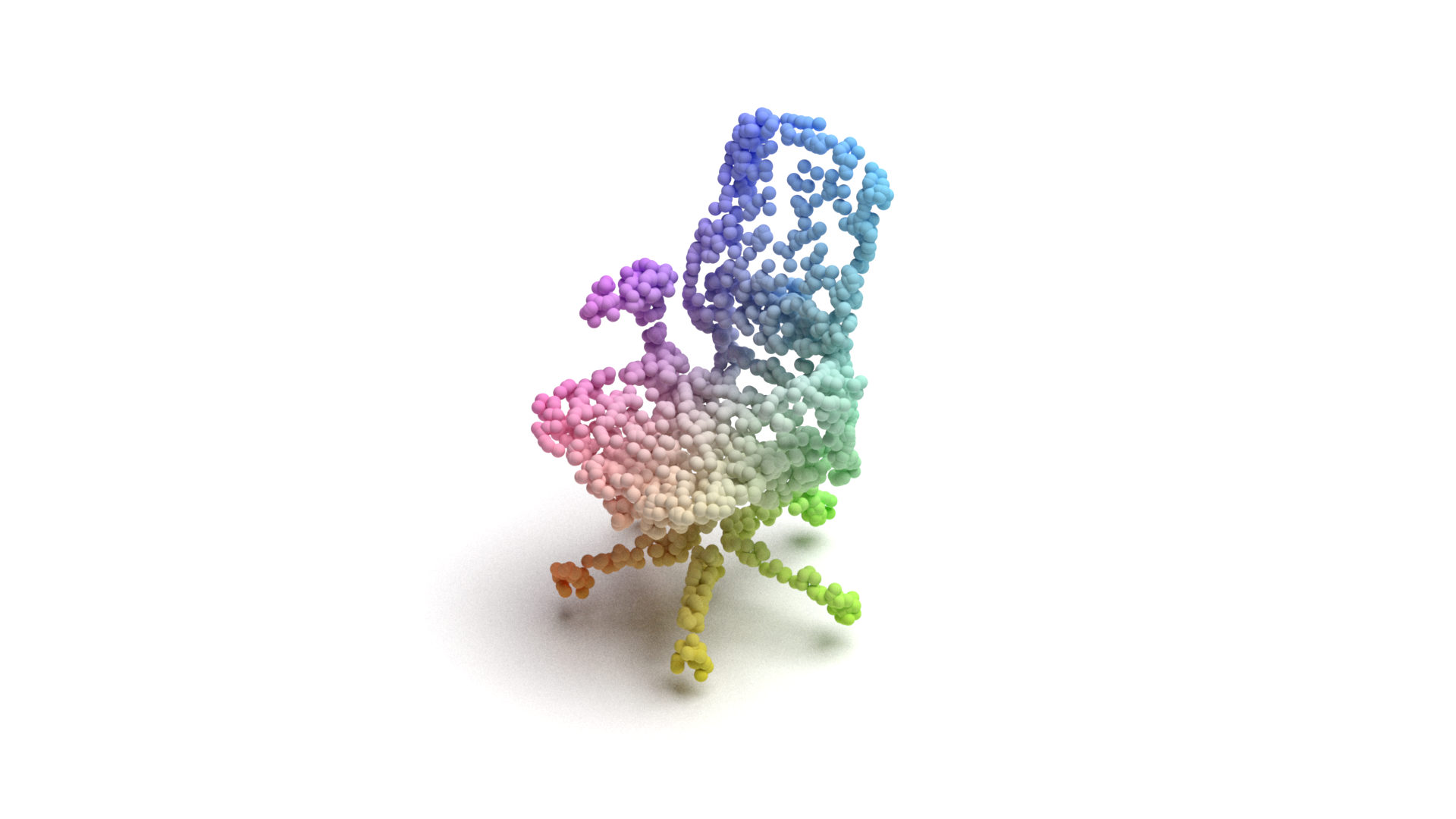}}           & 20\%    \\
\textit{Chair 70} & \includegraphics[width=2.5cm]{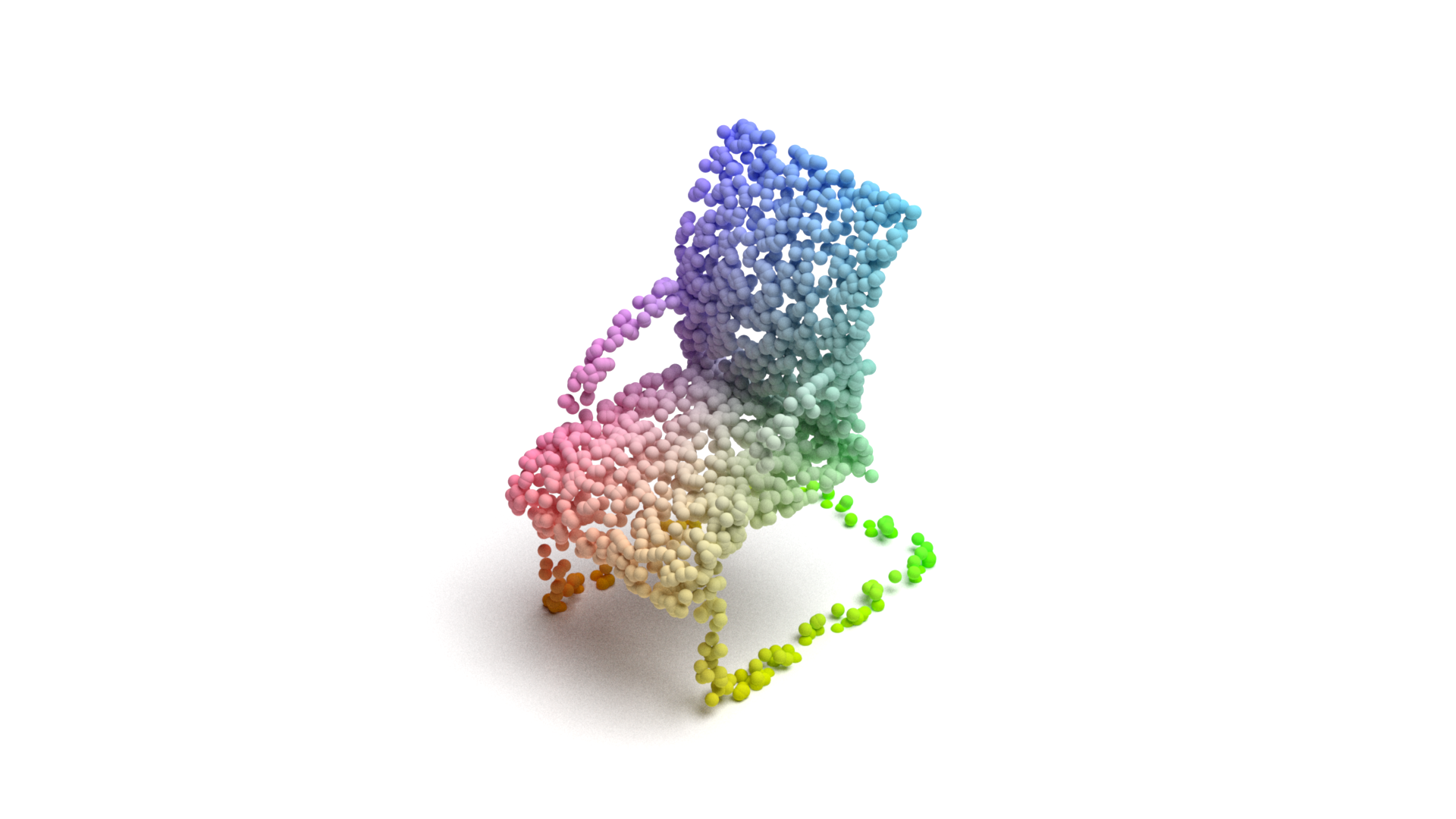}                &    20\%                       \\
\textit{Chair 87} & \includegraphics[width=2.5cm]{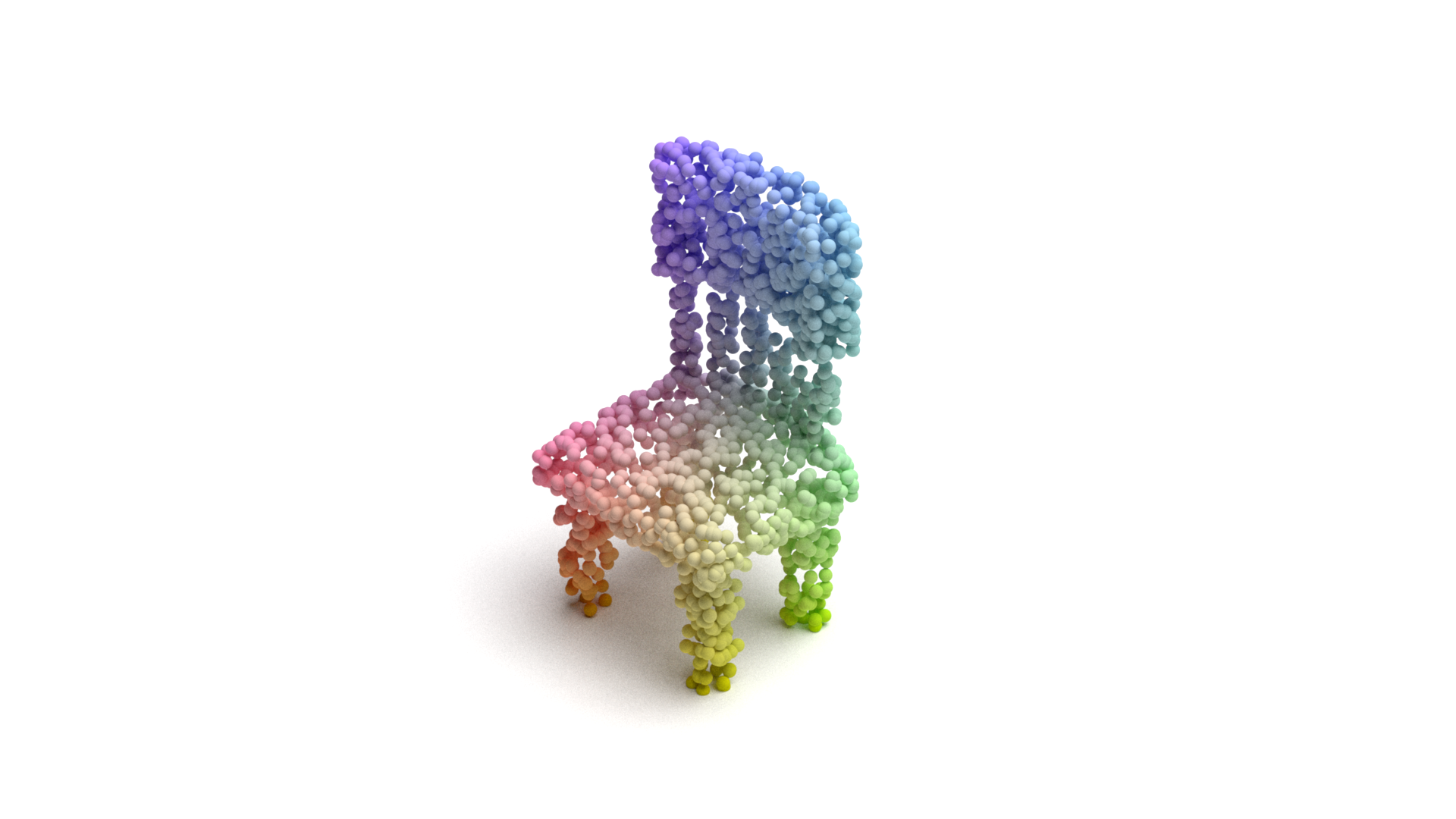}                &     20\%                     
\end{tabular}
\caption{Worst 3 chairs per accuracy}
\end{table}

\begin{figure*}[htbp]
    \centering
    \begin{subfigure}[b]{0.33\textwidth}
        \centering
        \includegraphics[width=\textwidth, trim=10cm 10cm 10cm 10cm, clip]{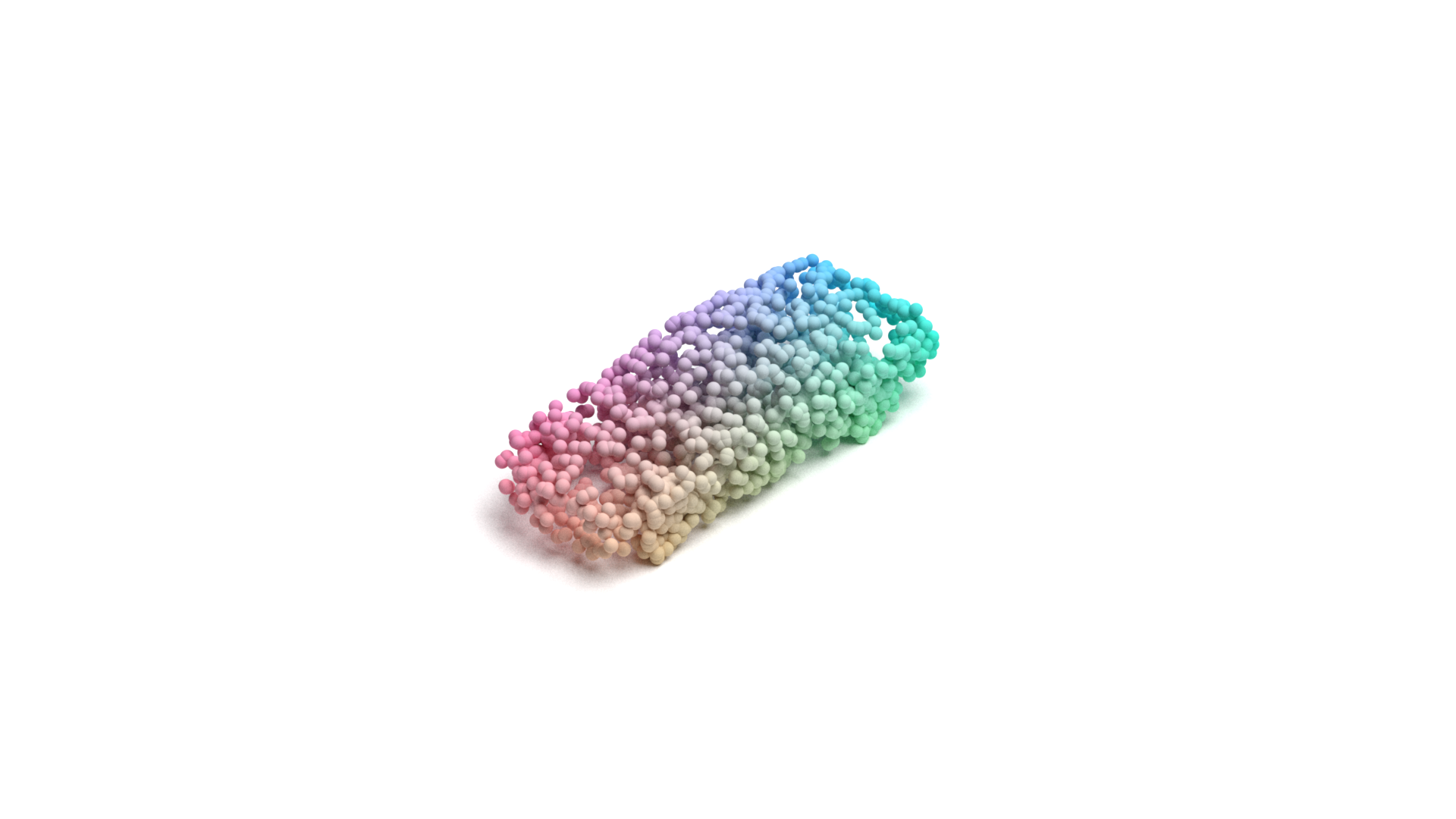}
        \caption{\footnotesize \textbf{Car \#184} \\ \footnotesize Acc: 30\%}
    \end{subfigure}
    \hfill
    \begin{subfigure}[b]{0.33\textwidth}
        \centering
        \includegraphics[width=\textwidth, trim=10cm 10cm 10cm 10cm, clip]{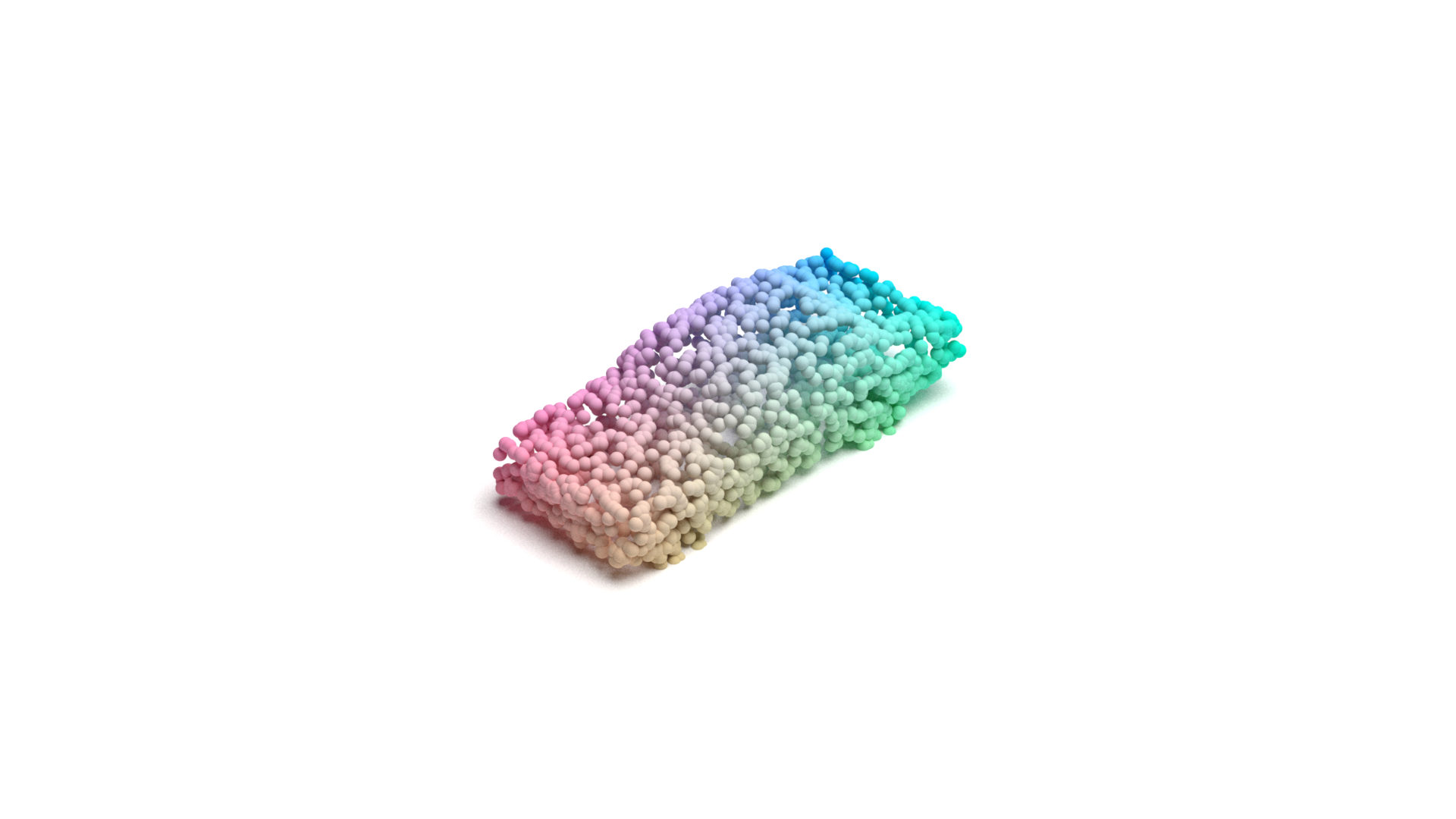}
        \caption{\footnotesize \textbf{Car \#39} \\ \footnotesize Acc: 28\%}
    \end{subfigure}
    \hfill
    \begin{subfigure}[b]{0.33\textwidth}
        \centering
        \includegraphics[width=\textwidth, trim=10cm 10cm 10cm 10cm, clip]{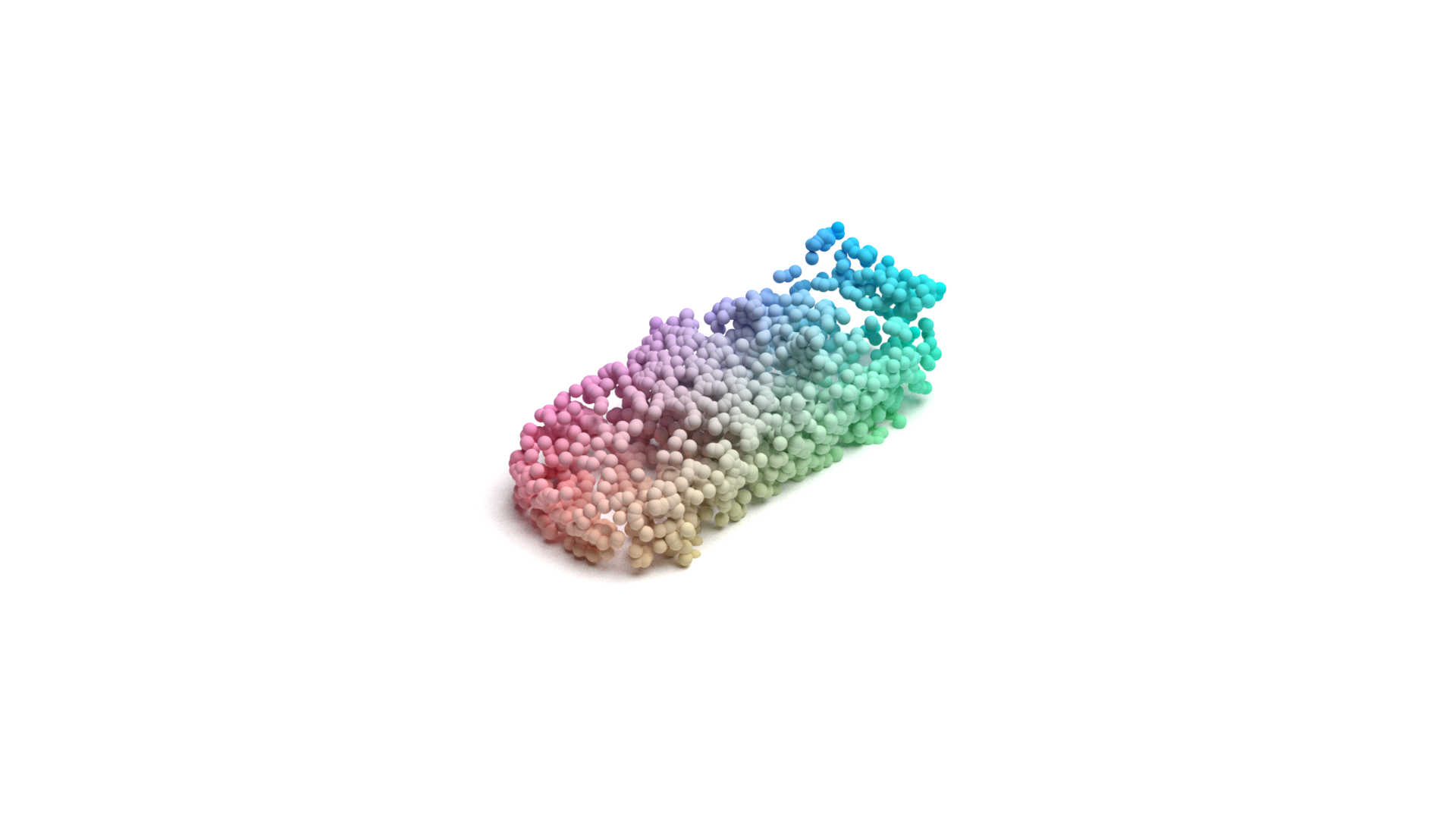}
        \caption{\footnotesize \textbf{Car \#142} \\ \footnotesize Acc: 27\%}
    \end{subfigure}
    
    \vspace{0.5cm} 

    \begin{subfigure}[b]{0.33\textwidth}
        \centering
        \includegraphics[width=\textwidth, trim=10cm 10cm 10cm 10cm, clip]{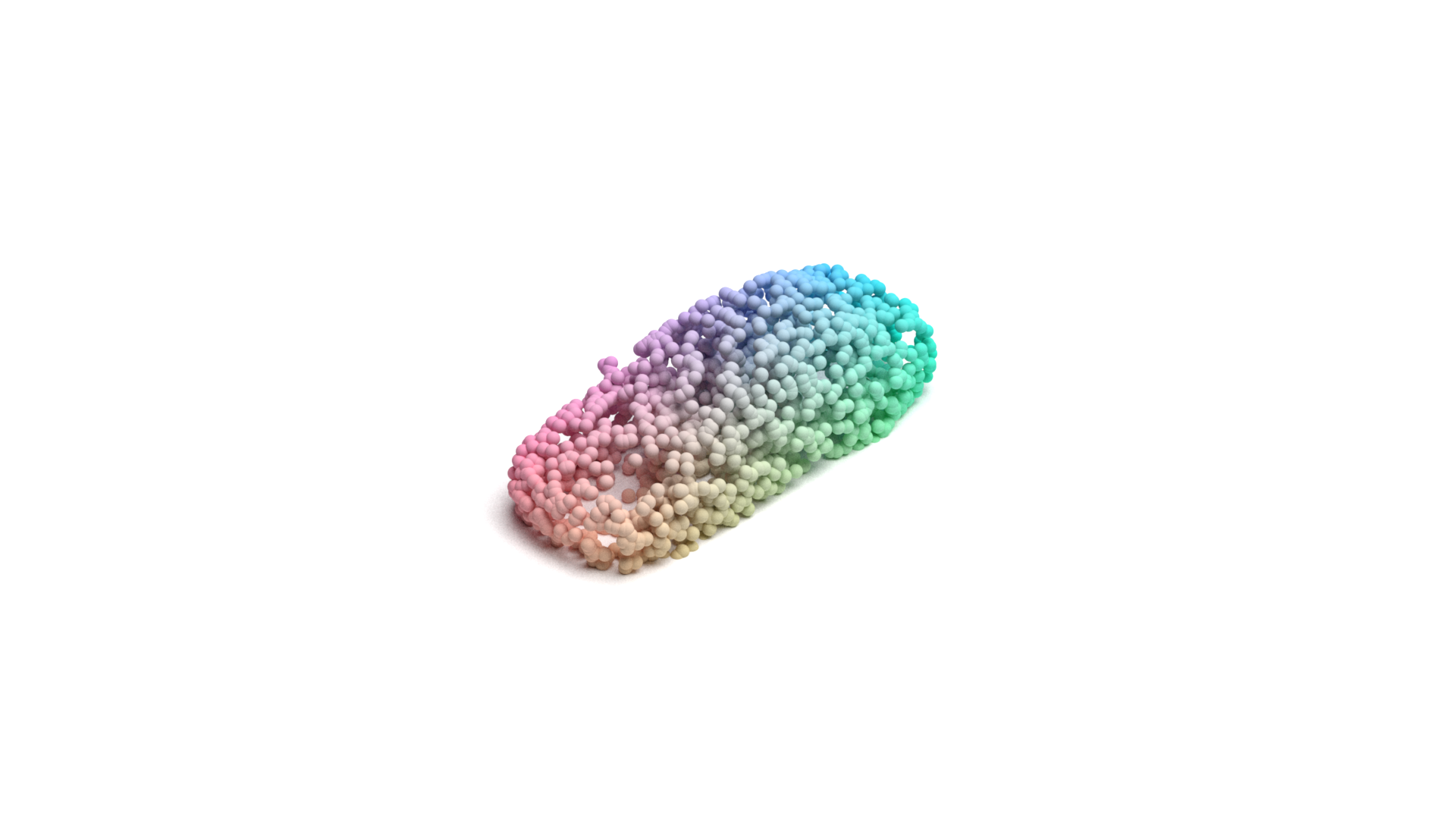}
        \caption{\footnotesize \textbf{Car \#149} \\ \footnotesize Acc: 19\%}
    \end{subfigure}
    \hfill
    \begin{subfigure}[b]{0.33\textwidth}
        \centering
        \includegraphics[width=\textwidth, trim=10cm 10cm 10cm 10cm, clip]{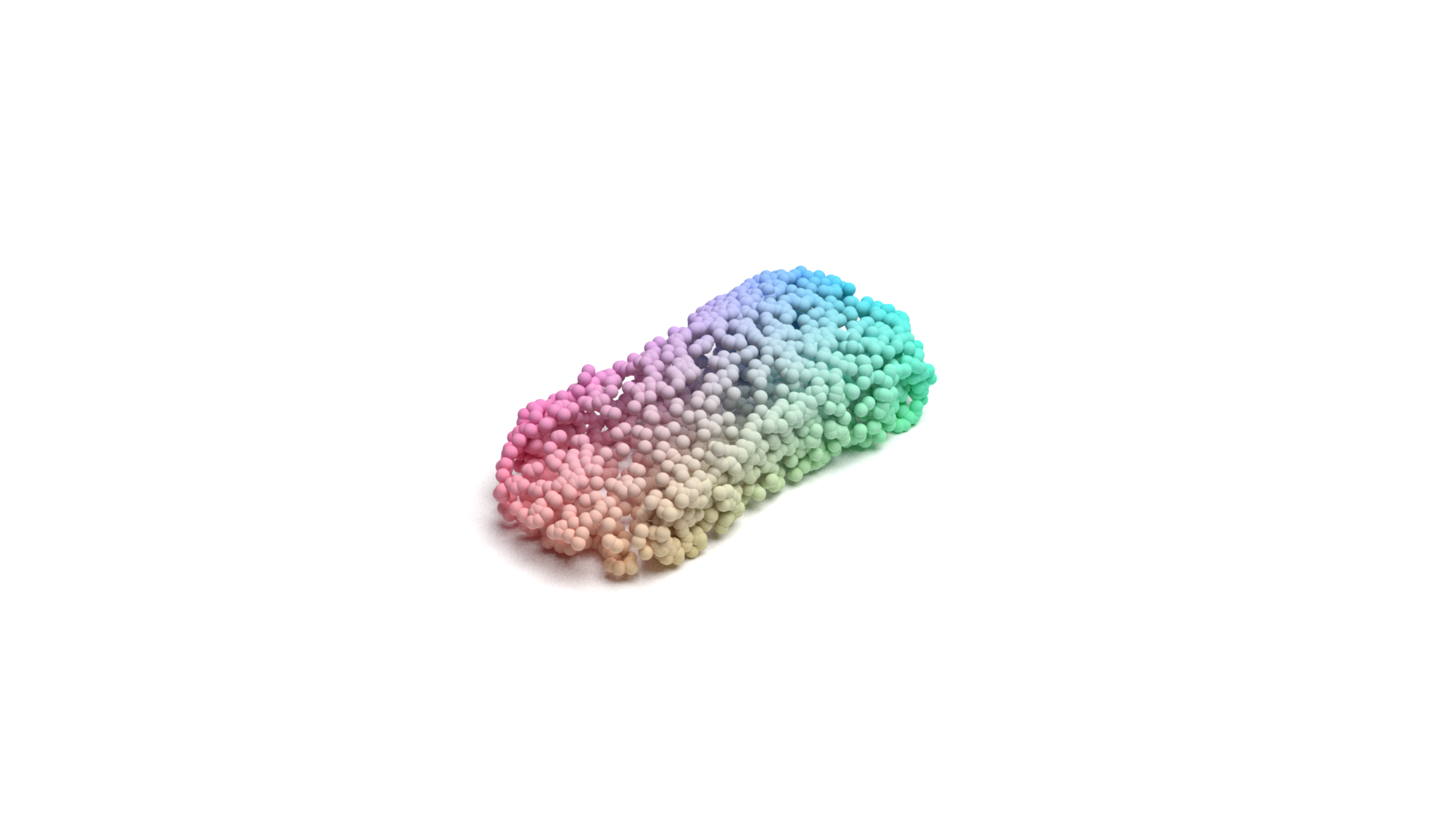}
        \caption{\footnotesize \textbf{Car \#152} \\ \footnotesize Acc: 20\%}
    \end{subfigure}
    \hfill
    \begin{subfigure}[b]{0.33\textwidth}
        \centering
        \includegraphics[width=\textwidth, trim=10cm 10cm 10cm 10cm, clip]{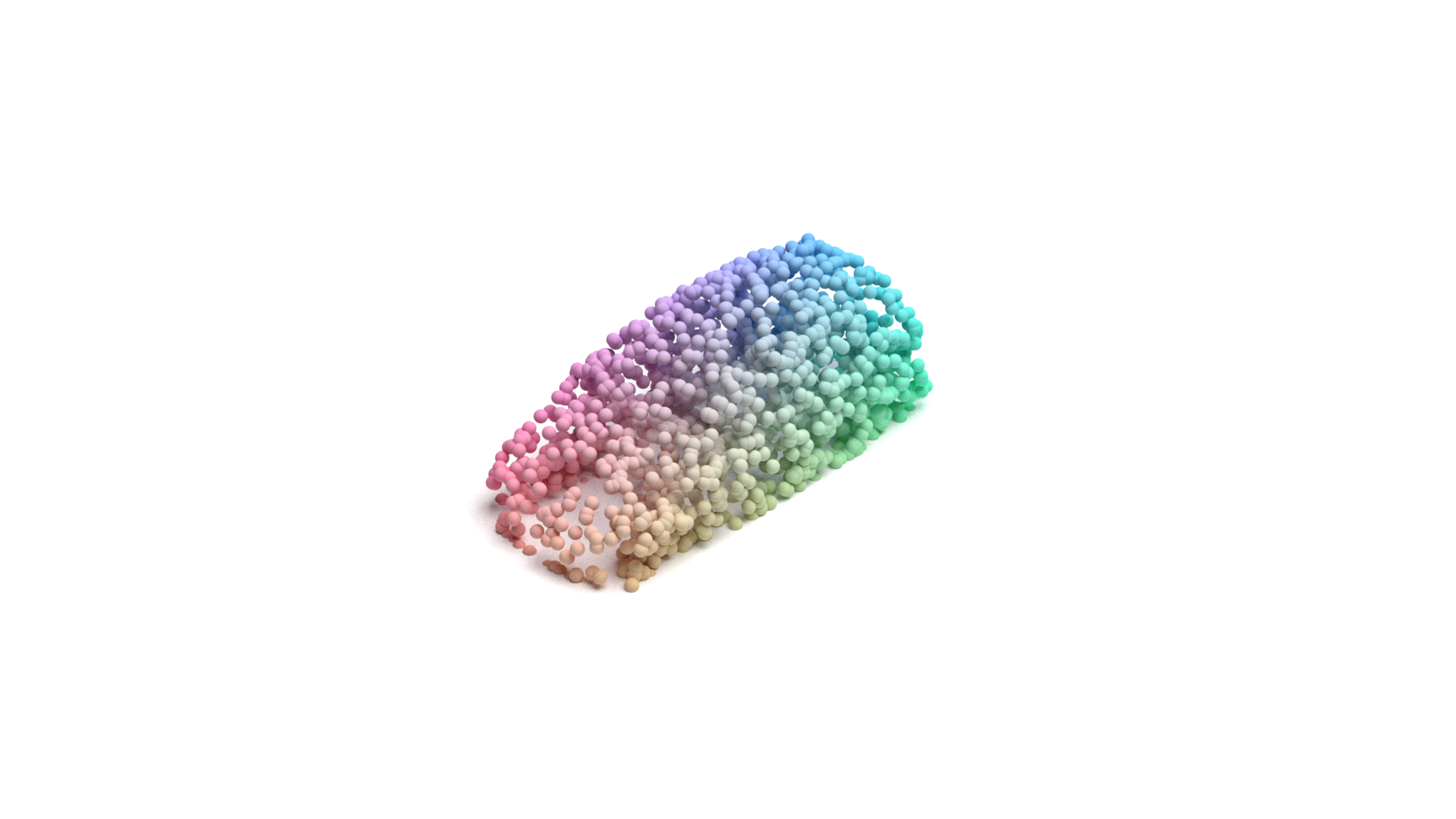}
        \caption{\footnotesize \textbf{Car \#60} \\ \footnotesize Acc: 19\%}
    \end{subfigure}
    
    \caption{\textbf{Visual Comparison}. comparison of best(a-c) and worst(d-f) performing cars with accuracy percentages}
    \label{fig:car_comparison}
\end{figure*}

\subsection{Ablation Studies}



To gain deeper insights into the contributions of different components in our model, we conducted ablation studies by systematically disabling specific features and quantifying their impact on performance.

For the MVDC model, let $S \times S$ denote the image size and $n$ the number of views. We observed that the diffusion process time $T$ increases non-linearly with both $S$ and $n$. Specifically, $T \propto S^2 \times n$, indicating that larger image sizes $S$ and an increased number of views $n$ result in significantly slower processing.

We ran inference at various image sizes to study its run time performance and relationship to the model performance. First, we confirmed that the inference time grows exponentially with larger image size, for a $512 \times 512$ resolution image and 500 sampling steps, the processing time was approximately 1.5 minutes per image, making it infeasible to evaluate at larger scale. Moreover, when we reduced the image size to $S=64 \times 64$ or $S=128 \times 128$, the classifier's performance degraded severely. The model exhibited a tendency to collapse, consistently predicting a single class $c$ regardless of the input views, suggesting that the classifier lost its ability to differentiate between classes under reduced image resolutions.

\begin{table}[h]
    \centering
    \caption{ \textbf{Ablation studies about image resolutions}. Inference time and accuracy analysis of MVDC model on 3 classes from shapenet: Airplane, Car, and Chair. The sample size is fixed at 200 steps.}
    \begin{tabular}{l|*{4}{c}}
        \toprule
        \textbf{Image} & \textbf{Inference} & \multicolumn{3}{c}{\textbf{Accuracy}} \\
        \cmidrule(lr){3-5}
        \textbf{Resolution} & \textbf{Time} & \textbf{Airplane} & \textbf{Car} & \textbf{Chair}  \\
        \midrule
        $64 \times 64$ & 1h03m  & 66.7\% & 64.8\% & 31.5\% \\
        \midrule
        $128 \times 128$ & 2h13m & 33.7\% & 66.7\% & 67.0\% \\
        \midrule
        $256 \times 256$ & 7h05m & 99.3\% & 98.7\% & 99.\% \\
        \midrule
        \bottomrule
    \end{tabular}
\end{table}






\section{Limitations}
\label{limitations}
One of the primary limitations of our approach is the computational cost. The 3D diffusion process currently requires approximately $20$ minutes per object on a T4 GPU, making it a time-intensive task. Similarly, the multi-view approach, while effective, is also relatively slow due to the independent processing of each view.

Regarding the Multi-View Diffusion Classifier, a significant limitation is that the views are processed individually and then aggregated through a majority vote, rather than being combined into a global latent vector as in the approach used by MVCNN \cite{su15mvcnn}. This method of independent view processing may not fully capture the holistic structure of 3D shapes, which could be better represented through a more integrated multi-view approach.

Due to time and computational constraints, we limited our experiments to 200 shapes per category. With access to more powerful GPUs and additional resources, future work could extend these experiments to a larger number of objects, potentially providing more comprehensive results.

\section{Discussion and Future Work}

The high classification accuracy on ID data indicates that the model effectively captures the distinguishing features of various 3D objects. 


The hierarchical latent space of LION played a crucial role in accurately representing both global and local features of 3D shapes, contributing to the model's overall performance compared the multi-view (see Section \ref{limitations} for more details). The diffusion process further enhanced the model's ability to denoise and classify complex 3D structures, providing a reliable mechanism for zero-shot classification.

These results highlight the potential of integrating generative models like LION with diffusion classifiers for advanced 3D shape analysis and classification tasks, particularly in scenarios involving diverse and unseen data. In fact, in this work, we delved into 3D diffusion models and present our method that enables zero-shot classification of 3D shapes in a robust manner. For future works, we wish to explore 3D diffusion capabilities in \textit{state-of-the-art} multimodal methods such as ULIP-2 \cite{Ulip_2_Xue2024}, integrated with PointBERT \cite{pointbert} architectures similar to the concurrent work \cite{DiffCLIP_Shen_2024_WACV}. We believe this will enhance the performance of these architectures and make them capable of 3D understanding.

\section{Conclusion}
In this paper, we propose a model that seamlessly integrates LION \cite{lion} with a diffusion classifier \cite{diffusion_classifier} to achieve accurate classification of 3D cars and chairs. The model’s success is driven by the hierarchical latent space and diffusion process, which together enable precise representation and classification of complex 3D shapes from the ShapeNet dataset \cite{chang2015shapenet}. Our approach, named \textbf{DC3DO}, demonstrates a 12.5\% improvement on average compared to multi-view methods, highlighting the potential of generative models in 3D object classification. This work suggests that future research could adapt generative models to discriminative tasks, potentially leading to enhanced classification and regression performance.

\section{Acknowledgments}
We express our gratitude to Alexander C. Li and Le Xue for their insightful initial discussions. We also extend our thanks to the mentors and volunteers of the Summer Geometric Initiative (SGI) at MIT for their invaluable guidance. A special thank you to Professor Justin Solomon for his coordination and funding support. We are also grateful to Google Cloud for providing credits and sponsorship, and to Paul Guerrero for generously sharing the render file used in the previous SGI.

{
    \small
    \bibliographystyle{ieeenat_fullname}
    \bibliography{main}
}

\end{document}